\pdfoutput=1

\documentclass[11pt]{article}

\usepackage{booktabs}
\usepackage{enumitem}
\usepackage{balance}

\usepackage{ACL2023}
\usepackage{graphicx} 

\usepackage{multirow}

\usepackage{times}
\usepackage{latexsym}
\usepackage{amsmath,amsthm,amssymb, graphicx, multicol, array, mathtools}
\usepackage{xcolor} 

\usepackage[T1]{fontenc}

\usepackage[utf8]{inputenc}

\usepackage{microtype}

\usepackage{inconsolata}

\usepackage{xspace}
\usepackage{amsmath}
\usepackage{amsfonts}
\usepackage{url}
\usepackage{marvosym}
\usepackage{ifthen}
\theoremstyle{plain}

\newcommand{\chatoDisplayMode}[1]{#1}

\definecolor{MyRed}{rgb}{0.6,0.0,0.0} 
\definecolor{MyBlack}{rgb}{0.1,0.1,0.1} 
\newcommand{\inred}[1]{{\color{MyRed}\sf\textbf{\textsc{#1}}}}
\newcommand{\frameit}[2]{
  \begin{center}
  {\color{MyRed}
  \framebox[.9\columnwidth][l]{
    \begin{minipage}{.85\columnwidth}
    \inred{#1}: {\sf\color{MyBlack}#2}
    \end{minipage}
  }\\
  }
  \end{center}
}

\newcommand{\note}[2][]{\chatoDisplayMode{\def\@tmpsig{#1}\frameit{{\Pointinghand} Note}{#2\ifx \@tmpsig \@empty \else \mbox{ --\em #1}\fi}}}
\newcommand{\todo}[2][]{\chatoDisplayMode{\def\@tmpsig{#1}\frameit{{\Writinghand} To-do}{#2\ifx \@tmpsig \@empty \else \mbox{ --\em #1}\fi}}}

\newcommand{\abbrevStyle}[1]{#1}

\newcommand{\ie}{\abbrevStyle{i.e.}\xspace}
\newcommand{\eg}{\abbrevStyle{e.g.}\xspace}

\newcommand{\vs}{\abbrevStyle{vs.}\xspace}
\newcommand{\etc}{\abbrevStyle{etc.}\xspace}

\newcommand{\xhdr}[1]{\vspace{1.7mm}\noindent{{\bf #1.}}}

\newcommand{\textcite}[1]{\citeauthor{#1} \shortcite{#1}}

\newcommand{\hide}[1]{}

\makeatletter
\newcommand{\iffont}[2]{\ifthenelse{\equal{\f@family}{#1}}{#2}{}}
\makeatother

\iffont{ptm}{
  \usepackage{mathptmx}

  \DeclareSymbolFont{greek}{OML}{cmm}{m}{n}
  \DeclareMathSymbol{\alpha}{\mathalpha}{greek}{"0B}
  \DeclareMathSymbol{\beta}{\mathalpha}{greek}{"0C}
  \DeclareMathSymbol{\gamma}{\mathalpha}{greek}{"0D}
  \DeclareMathSymbol{\delta}{\mathalpha}{greek}{"0E}
  \DeclareMathSymbol{\epsilon}{\mathalpha}{greek}{"0F}
  \DeclareMathSymbol{\zeta}{\mathalpha}{greek}{"10}
  \DeclareMathSymbol{\eta}{\mathalpha}{greek}{"11}
  \DeclareMathSymbol{\theta}{\mathalpha}{greek}{"12}
  \DeclareMathSymbol{\iota}{\mathalpha}{greek}{"13}
  \DeclareMathSymbol{\kappa}{\mathalpha}{greek}{"14}
  \DeclareMathSymbol{\lambda}{\mathalpha}{greek}{"15}
  \DeclareMathSymbol{\mu}{\mathalpha}{greek}{"16}
  \DeclareMathSymbol{\nu}{\mathalpha}{greek}{"17}
  \DeclareMathSymbol{\xi}{\mathalpha}{greek}{"18}
  \DeclareMathSymbol{\pi}{\mathalpha}{greek}{"19}
  \DeclareMathSymbol{\rho}{\mathalpha}{greek}{"1A}
  \DeclareMathSymbol{\sigma}{\mathalpha}{greek}{"1B}
  \DeclareMathSymbol{\tau}{\mathalpha}{greek}{"1C}
  \DeclareMathSymbol{\upsilon}{\mathalpha}{greek}{"1D}
  \DeclareMathSymbol{\phi}{\mathalpha}{greek}{"1E}
  \DeclareMathSymbol{\chi}{\mathalpha}{greek}{"1F}
  \DeclareMathSymbol{\psi}{\mathalpha}{greek}{"20}
  \DeclareMathSymbol{\omega}{\mathalpha}{greek}{"21}
  \DeclareMathSymbol{\varepsilon}{\mathalpha}{greek}{"22}
  \DeclareMathSymbol{\vartheta}{\mathalpha}{greek}{"23}
  \DeclareMathSymbol{\varpi}{\mathalpha}{greek}{"24}
  \DeclareMathSymbol{\varrho}{\mathalpha}{greek}{"25}
  \DeclareMathSymbol{\varsigma}{\mathalpha}{greek}{"26}
  \DeclareMathSymbol{\varphi}{\mathalpha}{greek}{"27}
  \DeclareSymbolFont{otone}{OT1}{cmr}{m}{n}
  \DeclareMathSymbol{\Gamma}{\mathalpha}{otone}{0}
  \DeclareMathSymbol{\Delta}{\mathalpha}{otone}{1}
  \DeclareMathSymbol{\Theta}{\mathalpha}{otone}{2}
  \DeclareMathSymbol{\Lambda}{\mathalpha}{otone}{3}
  \DeclareMathSymbol{\Xi}{\mathalpha}{otone}{4}
  \DeclareMathSymbol{\Pi}{\mathalpha}{otone}{5}
  \DeclareMathSymbol{\Sigma}{\mathalpha}{otone}{6}
  \DeclareMathSymbol{\Upsilon}{\mathalpha}{otone}{7}
  \DeclareMathSymbol{\Phi}{\mathalpha}{otone}{8}
  \DeclareMathSymbol{\Psi}{\mathalpha}{otone}{9}
  \DeclareMathSymbol{\Omega}{\mathalpha}{otone}{10}
  \DeclareSymbolFont{syms}{OML}{cmm}{m}{it}
  \DeclareMathSymbol{\partial}{\mathord}{syms}{"40}
  \DeclareMathAlphabet{\mathbold}{OML}{cmm}{b}{it}
  \DeclareSymbolFont{largesymbols}{OMX}{cmex}{m}{n}

}
\newcommand{\ignore}[1]{}

\title{Generating Faithful Synthetic Data with Large Language Models: \\ A Case Study in Computational Social Science}

\author{Veniamin Veselovsky{\textsuperscript{$\dagger$}},
Manoel Horta Ribeiro{\textsuperscript{$\dagger$}},
 Akhil Arora{\textsuperscript{$\dagger$}},\\
{\bf Martin Josifoski\textsuperscript{$\dagger$}},
{\bf Ashton Anderson\textsuperscript{$*$}}, 
{\bf Robert West\textsuperscript{$\dagger$}}\\
    \textsuperscript{$\dagger$} EPFL \textsuperscript{$*$}University of Toronto \\ 
    \texttt{firstname.lastnames@epfl.ch},
    \texttt{ashton@cs.toronto.edu}
    }

\begin{document}
\maketitle

\begin{abstract}
 Large Language Models (LLMs) have democratized synthetic data generation, which in turn has the potential to simplify and broaden a wide gamut of NLP tasks.
 Here, we tackle a pervasive problem in synthetic data generation: its generative distribution often differs from the distribution of real-world data researchers care about (in other words, it is \emph{unfaithful}).
 In a case study on sarcasm detection, we study three strategies to increase the faithfulness of synthetic data: grounding, filtering, and taxonomy-based generation.
We evaluate these strategies using the performance of classifiers trained with generated synthetic data on real-world data. 
While all three strategies improve the performance of classifiers, we find that grounding works best for the task at hand.
As synthetic data generation plays an ever-increasing role in NLP research, we expect this work to be a stepping stone in improving its utility.
We conclude this paper with some recommendations on how to generate high(er)-fidelity synthetic data for specific tasks. 
\end{abstract}

\begin{figure*}[h]
    \centering
    \includegraphics[width=\linewidth]{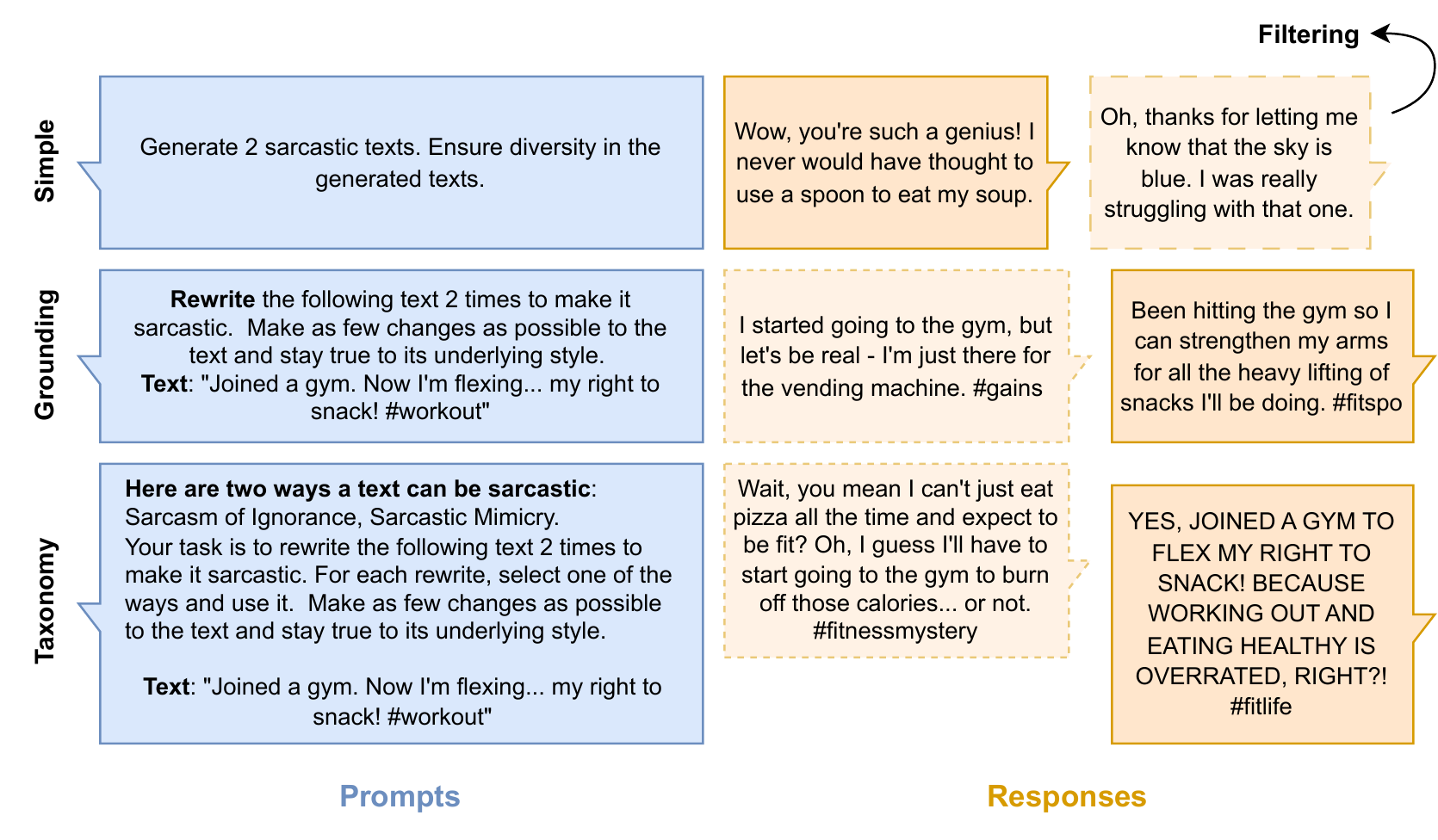}
    \caption{Depiction of the proposed strategies to increase the faithfulness of synthetically generated data. On the left-hand side, we depict different prompting strategies: asking an LLM to generate synthetic data with a simple prompt (\textbf{Simple)};
    grounding the synthetic data generation with real-world examples (\textbf{Grounding}-rewrite);
    and providing a taxonomy along with your prompt (\textbf{Taxonomy}). We also train a discriminator to distinguish between real and fake prompts and filter the data (as indicated by the dotted orange boxes on the right-hand side; \textbf{Filtering}).
    }
    \label{fig:motivation}
\end{figure*}
\section{Introduction}

From data annotation~\cite{gilardi2023chatgpt} to dataset creation~\cite{josifoski2023exploiting}, synthetic data offers previously unseen flexibility in the models we train~\cite{eldan2023tinystories} and in defining what and how we study the world around us~\cite{ziems2023can}. 
Further, large language models (hereinafter LLMs) are now easily accessible through APIs, substantially decreasing the expertise and the time necessary to generate synthetic data and labels.


Here, we examine a pervasive problem in synthetic data generation with LLMs: \textit{faithfulness}.
The generative distribution of synthetic data created by LLMs often differs from the distribution of real-world data that we care about~\cite{alaa2022faithful}. 
For instance, if we ask LLMs to generate tweets, these will likely be much better written than real tweets, and the topics and themes of those are likely to be less diverse.
This is problematic, as classifiers trained on synthetic data would be systematically biased and may not perform well in real-world contexts.

We study three strategies to increase the faithfulness of synthetic data generated by LLMs: grounding, filtering, and taxonomy-based generation. 
As illustrated in Fig.~\ref{fig:motivation}, grounding consists of providing real-world examples from a training set in the LLM prompt; 
filtering consists of using a discriminator model (trained to distinguish real and synthetic data) to cull unfaithful synthetic data;
and taxonomy-based generation consists of including a taxonomy in the prompt to encourage diversity.

We evaluate the aforementioned proposed strategies with a case study in Computational Social Science (CSS), a multidisciplinary field where easily accessible synthetic data and labels may be transformative in the years to come~\cite{bail2023can}.
Research in CSS often uses simple classifiers to estimate a linguistic characteristic or trait (referred to in the paper as a construct) in large text corpora, often obtained from the Web~\cite{salganik2019bit}. 
In that context, LLMs have been used to directly annotate the data in zero-shot fashion~\cite{ziems2023can}, and, more relevant to the work at hand, to create synthetic data to train models in complex or low-resource tasks~\cite{moller2023prompt}.

In the latter context, we consider the task of sarcasm  detection, and using an existing dataset evaluate the performance of each of the proposed strategies in increasing the faithfulness of synthetically generated data. 
Using the macro-F1 of the classifiers trained with different prompting strategies as a proxy for the faithfulness of synthetic data, we find that grounding provides the best performance our of all classifiers trained with synthetic data.
However, the model still performs worse in terms of macro-F1 than zero-shot ChatGPT annotation and a model trained on the real data.

\ignore{
\begin{figure*}[!htb]
\centering
\scriptsize
\begin{tabular}{|p{2cm}|p{8cm}|p{8cm}|}
\toprule
\textbf{Prompting Strategy} & \textbf{Prompt} & \textbf{Synthetically Generated Text} \\
\midrule
& Here are four ways a text can be sarcastic:
1. "Verbal Irony: Saying something but meaning the exact opposite.",
2. "Sarcastic Mimicry: Imitating or repeating others' statements sarcastically.",
3. "Sarcasm of Ignorance: Pretending to ignore or not understand the obvious."
4. Caps Lock Sarcasm: Using all caps in written communication for sarcastic emphasis.
Your task is to rewrite the following text 4 times to make it sarcastic.
For each rewrite, select one of the ways and use it. 
Make as few changes as possible to the text and stay true to its underlying style. 
Text: "Joined a gym. Now I'm flexing... my right to snack! \#workout" & \\
\bottomrule
\end{tabular}
\caption{Prompting strategies}
\label{fig:motivation}
\end{figure*}
}

\section{Related work}

\xhdr{Data augmentation}
In low-resource and unbalanced settings, augmenting datasets with synthetic data can improve model performance in a variety of NLP tasks,
including relation extraction~\cite{papanikolaou2020dare}, sarcasm detection~\cite{abaskohi2022utnlp}, translation~\cite{sennrich2015improving}, and sentiment analysis~\cite{maqsud2015synthetic}; see \citet{feng2021survey} for a comprehensive survey.
Specifically relevant to this paper is the work of \citet{moller2023prompt}, which uses ChatGPT to generate new samples for sentiment, hate speech, and a social dimension, a low-resource task. 
Finally,~\citet{anaby2020not} proposed a general methodology for fine-tuning a language model on small datasets. The authors highlight that the synthetic data was unfaithful to the real-world data distribution, thus warranting a filtering scheme to remove unfaithful data points.

\xhdr{Synthetic dataset creation}
Recent work has stretched beyond data augmentation to creating fully synthetic datasets. 
\citet{eldan2023tinystories} used LLMs to create ``Tiny Stories,'' showcasing how small a language model can  learn the language of 2 to 3-year-old children. 
This paper relied on a form of ``grounding'' to encourage diversity in the concepts discussed. 
Another work by \citet{josifoski2023exploiting} sampled knowledge graph triplets and generated texts using GPT-3. They then fine-tuned a model entirely on the synthetic data, and noted that the data was dissimilar from real human data. 

\xhdr{Synthetic data as a proxy for humans}
LLMs can also act as good proxies for specific human sub-populations~\cite{argyle2022out}, leading to a series of studies using LLMs as ``silicon samples''~\cite{argyle2022out,horton2023large,dillion2023can}. Typically, these analyses have been done through a variant of controlled text generation (review available here~\cite{zhang2022survey}). Further, an ever-increasing body of work illustrated the good performance of using LLMs as a proxy for human labeling~\cite{wang2023chatgpt,gilardi2023chatgpt,ziems2023can}. 

Naïve synthetic data generation with LLMs, \eg, the \textbf{Simple} strategy in Fig.~\ref{fig:motivation}, can lead to data that is unfaithful to the underlying real-world data distribution~\cite{josifoski2023exploiting}. 
This paper's contribution is to propose and evaluate prompting strategies that allow us to address this issue.

\section{Methods}

\subsection{Data}

We use the \emph{sarcasm detection} dataset from the SemEval-2022 Task 6~\cite{farha2022semeval}. 
The train set includes over two thousand self-disclosed instances of sarcasm being shared on Twitter. 
The reason we choose sarcasm is because it is an inherently difficult task to annotate, and construct to capture. Sarcastic texts are highly context-specific and ambiguous by nature. Annotating a sarcastic corpus has been a long standing problem, with sarcastic comments representing < 1\% of all text on social media (Reddit, for example). 
This renders it infeasible to blindly annotate texts since finding an instance of sarcasm is like searching for a needle in a haystack. 
Consequently, papers have traditionally relied on various heuristics to generate these datasets---like using the self-disclosed /s tag or asking users to share their own sarcastic Tweets (our task). These heuristics, however, lead to noisy labels and annotator bias~\cite{oprea2019isarcasm}.


\subsection{Evaluation} 
When evaluating how well our synthetic data captures a linguistic construct, we make the following assumption: \textit{if a construct is properly present in a synthetic dataset, then a model fine-tuned on that dataset will successfully generalize to a real human dataset.} We thus evaluate our synthetic data in three steps. 
First, we split human-annotated data into two groups train and test, throwing away the labels for our train data.
Second, we synthetically generate a new corpus through our various prompting strategies (see below).
Third, we fine-tune a model on the various generated synthetic datasets, and evaluate them on the test portion of the human-annotated data. 

\subsection{Prompting} 
To understand where synthetic data fails, we begin our analysis by manually inspecting the generated data.  
Three co-authors reviewed hundreds of examples of synthetically generated \vs real sarcastic texts and annotated their differences. We found that synthetic data generated with simple prompts:
1) exhibits a lack of topical diversity, \ie, it centered around a few topics of discussion;
2) lacks diversity in the construct of interest (namely sarcasm\footnote{There are many ways a linguistic construct like sarcasm can manifest (irony, over- or under-statement, satire, \etc), and typically the language model would retreat to superficial notions of sarcasm like beginning sentences with ``Oh'' or ``Wow''.}); and
3) are not well stylistically aligned with real data; authors could easily discriminate between synthetic and real texts. These three assumptions and corresponding prompt designs are described in Table~\ref{tab:method_explanation}.\footnote{The prompts in entirety are available at~\url{https://github.com/epfl-dlab/faithful-data-gen}.}

\begin{figure}[t!]
    \centering
    \includegraphics[width = 2in]{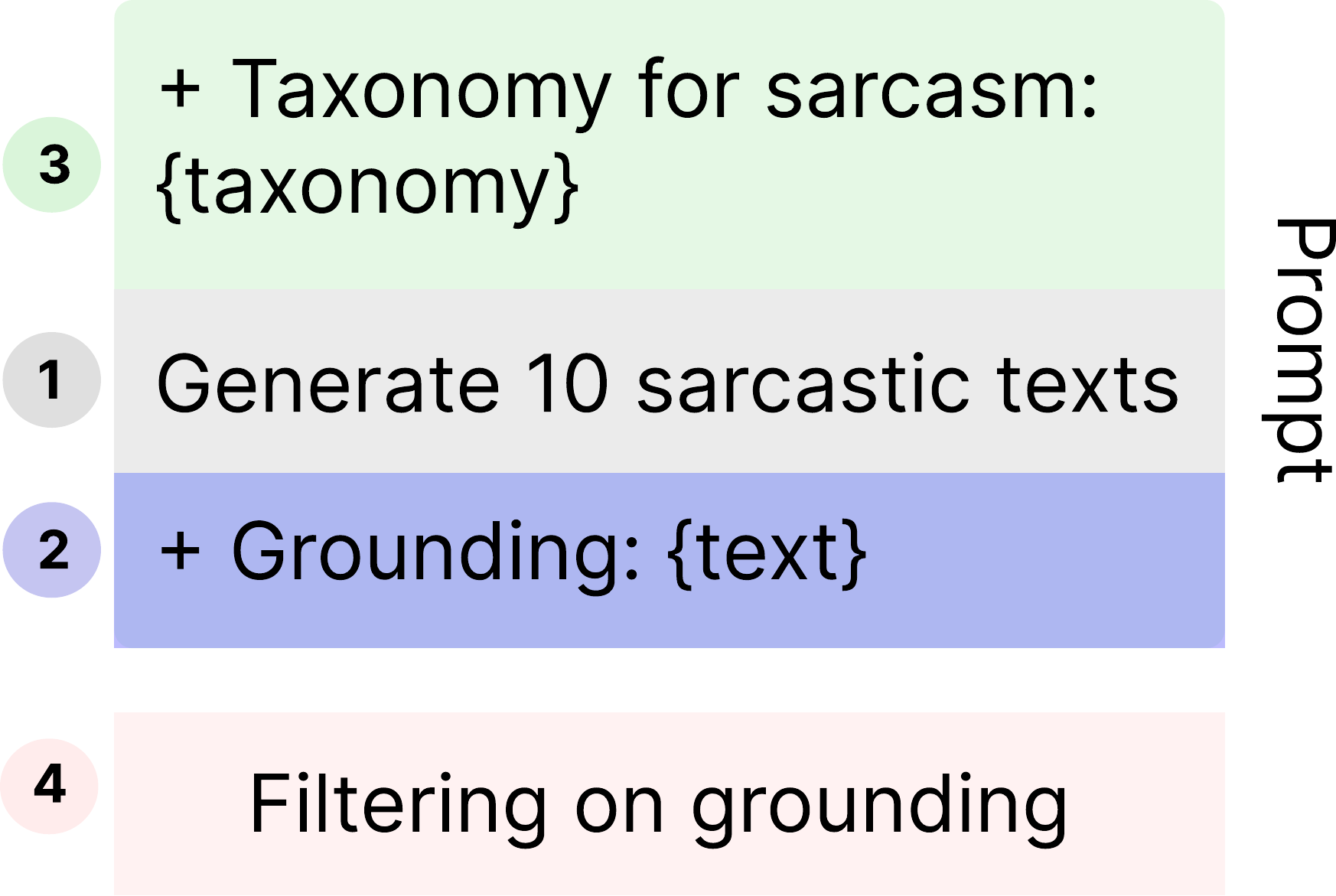}
    \caption{Our prompting approach consists of four modular steps. (1) Initiate the model to generate an initial set of 10 data points. (2) Apply a grounding technique as the model generates these 10 data points. (3) Further augment the grounding process by providing the model with an initial taxonomy. (4) Lastly, the results from the grounding phase are filtered through a real-synthetic classifier to ensure their authenticity.}
    \label{fig:prompting}
\end{figure}

We propose three prompting strategies to account for these limitations, each building off the next. 
Examples of how the prompts build off each other are illustrated in Figure \ref{fig:prompting} and discussed below. 

\begin{table}[t]
    \small
    \centering
    \begin{tabular}{ll}
         \multicolumn{1}{c}{\textbf{Goal}} & \multicolumn{1}{c}{\textbf{Strategy}}   \\
         \toprule
         Diversity in construct & Taxonomy creation \\
         \midrule
         Diversity in topics & Grounding \\
         \midrule
         Stylistic matching & Rewrite \\ \bottomrule
    \end{tabular}
    \caption{Description of objectives in synthetic data generation alongside specific strategies to achieve them.}
    \label{tab:method_explanation}
\end{table}

\xhdr{Grounding}  
We encourage topical diversity by \emph{grounding} the generations in real textual data. Specifically, in the prompt, we include an example of a real text and ask the model to either
1)~generate new semantically similar examples (like in~\citet{moller2023prompt} or \citet{eldan2023tinystories}) or 2)~rewrite the input text  (style transfer).

\xhdr{Taxonomy-based generation} 
We break up generation into two steps, asking the LLM to
1)~theorize $k$-ways a text can possess a specific construct and then sample across these $k$ approaches, and 
2)~rewrite the text according to a specific variant of the construct. 
The idea here is that generation based on an initial taxonomy can cover a wider segment of \emph{how} a text can actually convey a construct, improving the downstream model.

\xhdr{Filtering}
We fine-tune a model to discriminate between real and synthetic text and run that on the full batch of synthetically generated samples from the {\bf{Grounding}} data. 
We then cull the examples that have a high likelihood of being synthetic. 
We do this because, at times, the synthetic data has artifacts that are always associated with a construct. 
Specifically, we fine-tune a BERT model to distinguish between the first decoding (\ie, if we generate $10$ sentences, we only take the first sentence) and the real text to include a specific construct. 

For simple prompts, we ask the LLM to generate sarcastic and not-sarcastic text, and for prompts using \textbf{grounding}, we polarize each point in our dataset into two directions, \ie, making it both sarcastic and not-sarcastic.
In practice, this means that for each prompt in Fig.~\ref{fig:motivation}, we have an alternate version where we substitute the word ``sarcastic'' for ``not-sarcastic'', resulting in a  synthetic dataset that is balanced across the two classes.

\subsection{Models}
\xhdr{Generative model} To generate the synthetic data, we used ChatGPT.\footnote{\url{https://openai.com/blog/chatgpt}} 
The generation parameters for the model were set to \texttt{temperature: 1}, 
\texttt{top p: 1}, 
\texttt{frequency penalty: 0.5}, 
\texttt{presence penalty: 0.4}, 
\texttt{max tokens: 700}. 
We chose these parameters to maximize the diversity in the decoded text. 
The frequency penalty reduces the probability of a word depending on the frequency that it occurs, while the presence penalty puts a flat cost on each word when it occurs in the text. 
These two forms of penalties help encourage the model to produce higher perplexity texts instead of selecting the next most probable word. 
Moreover, temperature scaling produces a more shallow distribution over the next tokens, and a top-$p$ set to 1 will cause us to consider all these tokens. 

The generative data is then processed by removing artifacts of the generation. 
We defined these rules based on manual examination. The two most common problems that occurred were the model responding to the request in the affirmative (``Sure, here you go:'') and outlining which taxonomy it uses prior to generating the sentence (only present in the taxonomy generation prompting).
Both of these issues were addressed by splitting the first colon character ``:'' and restricting to text after it. 

\xhdr{Fine-tuned model} 
Similar to previous work, we fine-tune a \texttt{E5-base} model on the synthetic data~\cite{moller2023prompt,wang2022text}. 
This model was originally trained using a contrastive loss and achieves strong performance in a fine-tuned classification setting. 
During fine-tuning, we kept the settings from previous work with a learning rate of $2e^{-5}$, batch size of 32, and trained for 10 epochs.




\begin{table*}[t]
    \centering
    \begin{tabular}{l|c|c|c}
    \toprule
    \multirow{2}{*}{Prompting Strategy} & \multicolumn{3}{c}{\textbf{\emph{Sarcasm}}}  \\
     & Accuracy & Macro-F1 & Believability\\
    \midrule
    Simple & \textbf{0.71} & 0.48 & 0.04\\
    \hline
    Grounding & 0.67 & \textbf{0.55} & 0.13 \\ 
    \hline
    Grounding (rewrite) & 0.70 & \textbf{0.55} & 0.15 \\
    \hline
    Grounding + Taxonomy  & 0.67 & 0.51 & 0.20 \\
    \hline
    Grounding + Filtering & 0.27 & 0.26  & \textbf{0.56} \\
    \hline
    \hline
    Groundtruth annotations & 0.72 & 0.60 & 0.95 \\
    \hline
    All non-sarcastic & 0.77 & 0.43 & --- \\
    \hline
    Zero-shot ChatGPT &0.60 & 0.59 & --- \\
    \bottomrule
    \end{tabular}
    \caption{For different prompting strategies (rows 2 to 6) and baselines (rows 7 to 10), we show the accuracy, macro-F1 score, and believability in a held-out test set.}
    \label{tab:model_performance}

\end{table*}

\section{Results}

\xhdr{Model performance} 
We show the accuracy and the macro-F1 score for the different prompting strategies in the second and third columns in Table~\ref{tab:model_performance}. 
A baseline predicting all data points in the training set as not-sarcastic (``All non-sarcastic'') yields an accuracy of 0.72 and a macro-F1 score of 0.43.
In practice, we find that models trained in all prompting strategies perform worse accuracy-wise than this baseline, and thus it is more meaningful to compare their macro-F1 score.

We find that the ``simple'' prompting strategy generalized the worse (macro-F1 score of $0.48$), perhaps due to the lack of topical and construct diversity in the synthetically generated data.
Note that here we  prompted the model to generate 10 random instances of sarcastic and non-sarcastic texts five hundred times.
The two synthetic datasets that performed best (macro-F1 score:  $0.55$) were derived from the ``grounding'' prompting strategy, where the prompt asked the LLM to, given an example, generate semantically similar text (``Grounding,'' the 2nd row) or re-write it (``Grounding (rewrite),'' the 3rd row).
Prompting with grounding and an LLM-generated taxonomy yielded a result between the ``simple'' and the ``grounding'' prompting strategies (``Grounding + Taxonomy,'' macro-F1 score:  $0.51$).
Last, grounding the prompt and then filtering responses that were classified as synthetic with our discriminator yielded poor results  (``Grounding + Filtering,'' macro-F1 score $0.26$).

Finally, we note that zero-shot ChatGPT actually yields a higher macro-F1 score ($0.60$) than smaller models trained with synthetically generated data.

\xhdr{Believability} 
For each synthetic dataset generated, we further estimate how effective they are at fooling a synthetic \vs real classifier (which we refer to as the dataset's believability). 
The discriminator model was trained on individual generations of sarcastic and non-sarcastic text and then fine-tuned to predict if a text is sarcastic or not. We report the fraction of each dataset predicted to be real by this classifier in the 4th row of Table~\ref{tab:model_performance}, ``Believability.''
Note that for the groundtruth annotations (which are all real), we obtain a score of 95\%, meaning that the model believes that 95\% of the text was considered to be real by the classifier.

The dataset with the highest ``believability'' is the one created using the grounding and filtering strategies (``Grounding + Filtering,'' believability $0.56$). However, this metric may not capture faithfulness accurately in this case, as the criteria used for filtering are the same as the ones used to calculate the ``believability'' of a dataset.
Thus, of the remaining strategies, the ``Grounding + Taxonomy'' strategy presents the highest performance (predicted real: $0.20$), suggesting that data aided by a taxonomy picks up on fewer artifacts.
Unsurprisingly, the ``Simple'' strategy performs the worst, (predicted real: $0.04$), which is aligned with our qualitative analysis of the data, where we noted that most data points contain superficial sarcastic indicators like ``Oh'', ``Wow'', and question marks (``?'').
Last, grounded approaches perform better than the simple strategy (predicted real: $0.13$ for ``Grounding'' and $0.15$ for ``Grounding (rewrite)''). 

\xhdr{Key takeaways}
Through the process of generating synthetic data, we drew  takeaways that can be beneficial for future studies using synthetically generated data for either augmentation or as the entire dataset. We list these findings here:

\begin{itemize}
    \item When producing synthetic data, it is necessary to generate several sentences for each individual real sample. Typically, the later generations capture more interesting forms of sarcasm than the initial generation and cover a broader range of topics.
    
    \item Grounding data is a key aspect of generating synthetic data. Without grounding, the model tends to generate texts that are specialized in terms of topics discussed and constructed used. 
    
    \item Taxonomy creation can be useful for making the data appear real. However, it performs worse than grounding at staying true to the underlying construct. One potential reason for this is that we assume a uniform distribution over subvariants of sarcasm. This assumption is unlikely to hold in practice---in real life, there are a few types that represent most forms of sarcasm, with the rest representing a long tail. Applying a prior to the types of sarcasm we are likely can lead to more realistic generations.
    
    \item Filtering works poorly. This result is surprising given its prevalence in other data augmentation studies. This may be improved through a better classifier. 
    
    
    \item A small capacity model like E5 may not be capable of capturing complex linguistic features like sarcasm. It may be a worthwhile effort to fine-tune on a larger model like \texttt{Flan-T5}.
\end{itemize}







\section{Discussion}

\subsection{Summary of findings}
Investigating the ability of LLMs to generate \emph{faithful} synthetic data, we find that simple prompting strategies result in data that lacks diversity and differs stylistically from real-world data. 
To address these issues, we propose a suite of improved prompting strategies, namely, `grounding,' `filtering,' and `taxonomy-based generation,' which we qualitatively find to generate samples that are more faithful to the real-world data distribution. 
Further, comparing the performance of classifiers trained with synthetic data generated using our proposed strategies on a downstream task of sarcasm detection, we find that `grounding' resulted in the highest improvement, thereby indicating the importance of closely capturing topical diversity for the considered tasks.

\subsection{Implications}
We argue that the implications of the aforementioned findings are three-fold. 

First, our results suggest that synthetic data generation can be a resource-friendly alternative to human annotation achieving results five macro-F1 points worse than zero-shot annotation and a model trained on the real data. 
With only a few examples of data of the kind researchers wish to study (\eg, sarcastic tweets), they could bootstrap a synthetic dataset that can be used to train relatively simple, yet effective and easily deployable models. 
This strategy could also alleviate privacy concerns associated with using real-world data, allowing the study of sensitive topics without relying on collecting data from contexts where personally identifiable information is present (\eg, social media). 

Second, synthetic data generation could be a helpful strategy for training future (potentially smaller) language models. 
Previous work has shown that smaller language models fine-tuned using well-curated samples from a constrained domain can outperform larger models on specific tasks~\cite{zhou2023lima}, and with our prompting strategies, this fine-tuning process could be bootstrapped with another language model, \ie, one could automatically generate this well-curated sample. 
More broadly, as language models scale up, and organizations require more and more data to train these models, synthetically generated data may be needed to continue the improvement of these models. Our work could be seen as a stepping stone for more research in this direction.

Finally, we hope that the proposed strategies enable more fine-grained analyses in fields like Computational Social Science that leverage NLP to study human-made constructs. Constructs like sarcasm are not black and white and reflect the subtle complexities of human language; sarcasm can really take many sub-forms like hyperbole, satire, irony, understatements, rhetorical questions, juxtaposition, and sardonic humor. 
Building a model to detect these classes of sarcasm can be intractable. 
Do we search for distinct datasets for each of these types of sarcasm? Do we annotate a large corpus of sarcastic texts to fit into this taxonomy? It’s not entirely clear. However, this could be done with the taxonomy-based prompting strategy proposed in this work.



\subsection{Limitations and Future Work} 

Owing to its superior efficiency and cost effectiveness, we used ChatGPT for generating synthetic data in this work. However, in the future we aim to repeat all the analyses using data generated via GPT-4, which has shown to achieve substantial improvements over ChatGPT~\cite{bubeck2023sparks}. In the same vein, we would like to fine-tune a larger language model on the order of hundred million parameters for the downstream task of sarcasm detection. This is primarily because sarcasm detection is a difficult task, and therefore could benefit from the abilities that only emerge in LLMs at scale~\cite{wei2022emergent}.

Next, we would also like to extend our analyses to diverse NLP tasks. While the present work showcases the ability of our proposed prompting strategies to generate more faithful synthetic data using the task of sarcasm detection, our strategies are general and can be applied for other NLP tasks. 

From an evaluation standpoint, we use the downstream performance of classifiers trained on the generated synthetic data to quantitatively assess the quality of generations. However, this is inherently a proxy for evaluating data faithfulness. In the future, we would like to perform a more direct evaluation, such as conducting a turing test, by asking humans to distinguish between real and synthetically generated data.

Finally, we intend to perform extensive tuning of different components of our pipeline. For example, while we fix the number of re-writes to 10, it would be fruitful to identify the optimal value of the number of re-writes as well as understand its relationship with the complexity of the underlying task. Similarly, following the success of self-refinement~\cite{madaan2023self}, we would like to explore the use of iterative refinement strategies to discriminate between real \vs synthetic data, which is currently performed in a single filtering step.

\section*{Ethical considerations} All the datasets and resources used in this work are publicly available and do not contain any private or sensitive information about individuals. Moreover, all the findings are based on analyses conducted at an aggregate-level, and thus, no individual-level inferences can be drawn. However, human-like synthetic data can be used maliciously. We acknowledge this concern.

\ignore{
Human data is inherently random. There are many ways to say one thing, and humans use them all. However, when this gets mapped over to an LLM setup, we tend to lack this diversity and observe the averaging of text---text sounds more and more low perplexity. Given the nature of LLMs, it is difficult to change this since it is inherent to their nature of being (next token prediction). Our work illustrates an example of how we can artificially induce diversity into a black box model. In the future, we believe it can be beneficial to explore how adjusting the parameters from these models.

In this paper, we add to the literature for synthetically generating social data and offer a general framework for generating human-like data. When generating human-like data, there are three areas that we need to stay true to. Here I will limit the scope of discussion to generating data for a specific task (construct present in text) that we are studying.
First, we require that the data is topically diverse (similar to human-data). When asking the model to generate texts, the texts generated are typically spread across few sets of topics that have a high probability of occurring in text. Second, we want to the data to have diversity in the use of a construct we aim to generate in the text. For example, there are many ways for a text to sarcastic, yet the model typically generates one form of sarcasm. Third, we need the data to match the underlying style present in the human-like data. For instance, if we are interested in studying Twitter, we want the text to be ``Tweetable''. These three factors should be present in the generated data.

We require no annotated data, unlike the Luca approach where they assumed the existence of these labels. \ie, for us 

A problem with the Luca approach is that they assume that output labels exist, whereas our approach doesn't require this. These labels might be intractable to get or unavailable.
}

\bibliography{00paper}
\balance
\bibliographystyle{acl_natbib}

\appendix

\end{document}